\documentclass{article}

\PassOptionsToPackage{numbers, compress}{natbib}

 \usepackage{graphicx}
 \usepackage{enumitem}
\usepackage[preprint]{neurips}

\usepackage[utf8]{inputenc} 
\usepackage[T1]{fontenc}    
\usepackage{hyperref}       
\usepackage{url}            
\usepackage{booktabs}       
\usepackage{amsfonts}       
\usepackage{amsmath}       
\usepackage{microtype}      
\usepackage{xcolor}         
\usepackage{ulem}
\usepackage{listings}
\usepackage{courier} 
\title{The Smart Buildings Challenge: Learning Foundation Models for Real-World Optimization at Scale}

\author{
Judah Goldfeder \\ Columbia University \\ {\tt jag2396@columbia.edu}
\And
Ayush Rai \\ Harvard University \\ {\tt ayush\_rai@seas.harvard.edu}
\And
Philippe M. Wyder \\  Distyl AI \\ {\tt  philippe@distyl.ai}
\And
Na Li \\ Harvard University \\ {\tt nali@seas.harvard.edu}
\And
Victoria Dean \\ Princeton University \\ {\tt 
vdean@princeton.edu}
\And
Donna Vakalis \\  Mila \\ {\tt donna.vakalis@mila.quebec}
\And
Jiong Lin \\  Columbia University \\ {\tt jl6017@columbia.edu}
\And
Bing Dong \\ Syracuse University \\ {\tt  bidong@syr.edu}
\And
Michael Rossetti \\  Google \\ {\tt mjrossetti@google.com}
\And
Nathan Kutz \\  Autodesk Research \\ {\tt kutz@uw.edu}
\And
Sohei Arisaka \\  Kajima \\ {\tt s.arisaka@kajima.com.sg}
\And
John Sipple \\  Google \\ {\tt sipple@google.com}
\And
Hod Lipson \\Columbia University\\ {\tt hod.lipson@columbia.edu}
\And
Yoshua Bengio \\  Mila \\ {\tt yoshua.bengio@mila.quebec }
}

\begin{document}

\maketitle
\begin{abstract}
Buildings cause nearly 40\% of carbon emissions in the US. Of total building emissions, HVAC systems are responsible for approximately 40-60\% and account for 15\% of global energy consumption. Despite advances in artificial intelligence, most buildings still operate as inefficiently as they did decades ago. Building optimization, perhaps more so than any other optimization problem, has enormous potential for Machine Learning (ML) driven impact on climate change, in both the short and long term. One of the biggest obstacles for building optimization is the lack of policy transfer. Most building control systems, whether simulation or learning based, don't generalize: they only work well for the building they were developed for, but not for other buildings. To solve this problem, we propose a challenge for the development of building foundation models that easily adapt to new buildings with minimal configuration and thereby enable building optimization to scale across the globe. Participants will build control policies on a set of simulated buildings, and have their policies evaluated on an unseen set of buildings. The buildings vary in size, layout, weather, material properties, and HVAC system. Successful outcomes from this initiative promise a massive leap forward in global sustainability efforts and demonstrate the great potential of ML to combat climate change.
\end{abstract}

\paragraph{Keywords} Foundation Models, Transfer Learning, Climate Change, Building Optimization.

\section{Competition description}
\subsection{Background and impact}

\begin{figure}
    \centering
    \includegraphics[width=\linewidth]{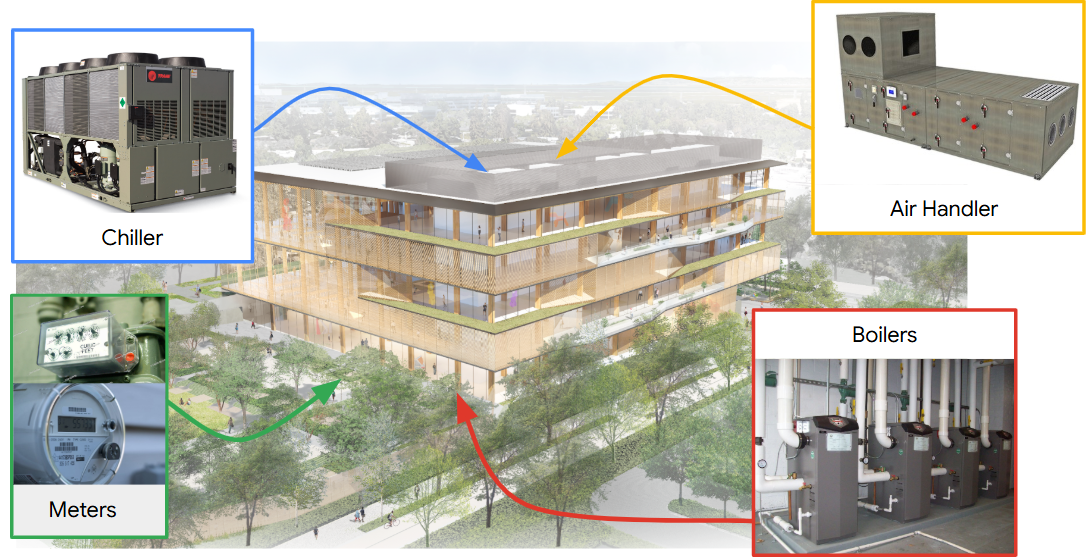}
    \caption{Illustration of an office building and its HVAC devices.}
    \label{fig:device_locations}
\end{figure}

Commercial buildings account for 17\% of total US carbon emissions \citep{eiaFrequentlyAsked}; within buildings, HVAC systems (see Fig.~\ref{fig:device_locations}) are responsible for roughly 40--60\% of energy use \citep{perez2008review}, or $\sim$15\% of global energy consumption \citep{asim2022sustainability}. Despite decades of active research \cite{fong2006hvac,trvcka2010overview,riederer2005matlab,park1985overview,trvcka2009co,husaunndee1997simbad,trcka2007comparison,blonsky2021ochre,crawley2001energyplus}, broad real-world adoption of optimized HVAC control remains elusive.

\paragraph{Current limitations.}
Most control policies---simulation or ML based---fail to generalize across buildings. The two primary reasons are \emph{system-specific dependencies} (solutions are often tied to a particular building management system, weather pattern, HVAC system, or data ontology, which prevents them from adapting to other systems) and \emph{lack of transfer training} (most controllers are trained on a single building because cross-building data is scarce and hard to come by).

\paragraph{Competition goal.}
A building foundation control model that could easily adapt to new buildings with minimal configuration would enable building optimization to scale to thousands of buildings with significantly less overhead---lowering the barrier of adoption. Recent evidence suggests that such a building model is within reach. Foundation models in other domains have demonstrated remarkable results in transfer learning, from vision\cite{awais2025foundation} and language\cite{myers2024foundation} to robotics\cite{firoozi2023foundation} and control\cite{yang2023foundation}, and progress is especially bolstered when there is community support behind a well adopted benchmark\cite{goldfeder2026driving,wyder2026common}, something we are hoping to foster within the building science and ML community (a successful recent example of this in the related ML and sustainability community is the AI City Challenge\cite{tang202610th}). This competition is trying to answer whether we can create a building foundation model that effectively transfers across many buildings. We will use a simulated environment for the contest, and will evaluate winning solutions on real buildings after the contest ends.

\paragraph{Impact.}
A scalable building model that unlocks even a 5\% reduction in carbon emissions could reduce the total carbon emissions of the US by roughly 1\%. This is especially true if steps are taken to ease public adoption\cite{goldfeder2024reducing}. Additionally, a building foundation model can lower costs of adopting optimal control mechanisms by removing the need for hard-to-calibrate simulations. Once adoption costs are lowered, widespread industry adoption can occur. We expect this competition to be of interest to the broader community, no domain knowledge will be needed to participate.

\subsection{Related Work}
HVAC control has garnered considerable interest\citep{fong2006hvac} in recent years \citep{kim2022energy}, and while alternatives like Model Predictive Control (MPC) exist\citep{taheri2022model}, a growing portion of the literature has considered Reinforcement Learning \citep{yu2021review,mason2019review,yu2020multi,gao2023comparative,wang2023comparison,vazquez2019reinforcement,zhang2019whole,fang2022deep,goldfeder2024reducing}. Of course, this requires a simulation and many works have explored this topic\citep{trvcka2010overview,riederer2005matlab,park1985overview,trvcka2009co,husaunndee1997simbad,trcka2007comparison,blonsky2021ochre,jiang2026bestopt}. EnergyPlus \citep{crawley2001energyplus}, a high-fidelity simulator developed by the Department of Energy, is commonly used \citep{wei2017deep,azuatalam2020reinforcement,zhao2015energyplus,wani2019control,basarkar2011modeling}, but suffers from scalability and configuration challenges.
To overcome the limitations of these methods, some work uses a hybrid approach \citep{zhao2021hybrid,balali2023energy,goldfeder2023lightweight,zhang2023bear,klanatsky2023grey,drgovna2021physics,jiang2025physics}, as does our simulator.

\subsection{Novelty}
To our knowledge, no previous competition has focused on foundation models for buildings, that is, on creating digital twin models that transfer from building to building. This is arguably the largest unsolved problem in this entire domain of research, and has been a major focus of recent work\cite{berkes2025hot,taboga2026building2building,raisch2025gentl,raisch2025adapting,krug2025builda}.

\subsection{Data}\label{sec:data}

The competition makes use of the HVAC simulation platform\cite{goldfeder2023lightweight,goldfeder2024real} from the Smart Buildings Control Suite\cite{goldfeder2025smartbuildingscontrolsuite}, along with some improvements from the open source community\cite{ham2026tensorized}. For the contest we created a random building generator that interfaces with the simulator. This allows for an infinite set of potential building simulators, each with different conditions. The generator varies the number of floors (2--4), width and length, perimeter shape, room layout, occupancy, envelope construction (six material properties), and the number of Air Handler Units (AHUs) and the zones they serve (at most 10 zones per AHU); the hot water system remains a single system per building. Each zone has its own VAV box. Each building is exposed as a Gym-compatible RL environment.

\begin{figure}[ht]
    \centering
    \includegraphics[width=\linewidth]{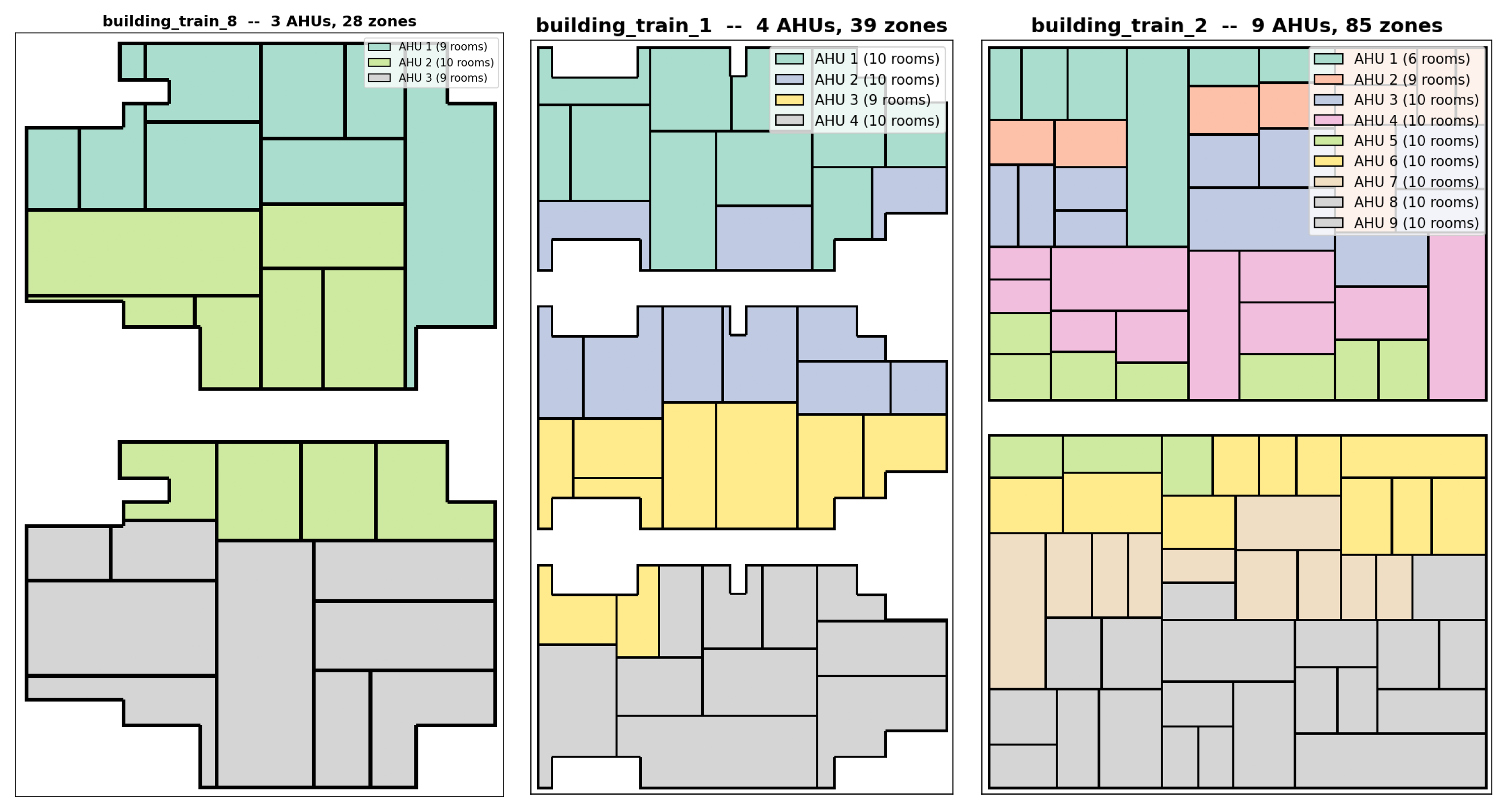}
    \caption{Three training buildings. Floor plan, zone count, and AHU-to-zone assignment differ per building (colors show the AHU serving each zone).}
    \label{fig:floorplan}
\end{figure}

\paragraph{Building split.}
Participants will be given 6 training buildings, each paired with one recorded weather strip from each of five climate stations (30 training scenarios). The agent will be evaluated on 6 unseen buildings from the same generator and parameter ranges, at 5 climate stations that appear nowhere in the training set, so the transfer is across both buildings and climates.

\paragraph{Availability and licensing.}
The simulated buildings and corresponding weather scenarios are generated procedurally by an open-source generator built on the Smart Buildings Control Suite~\cite{goldfeder2025smartbuildingscontrolsuite}. They contain no personally identifiable information, no real building telemetry, and no human subjects data, and therefore comply fully with the \href{https://neurips.cc/public/EthicsGuidelines}{NeurIPS Code of Ethics} on privacy, consent, and fair use. The generator, the 6 development buildings, and the training weather strips will be released under an open-source license alongside the starter kit at competition launch. The 6 held-out evaluation buildings and the 5 held-out stations are kept confidential by the Organizers and released publicly only after the competition concludes.

\paragraph{Scale and statistical power.}
Because the environment is procedural, the effective dataset is unbounded: participants may sample arbitrarily many training buildings, and Organizers can sample fresh evaluation buildings to defeat memorization. The default training pool of 6 buildings $\times$ 5 stations yields 30 development scenarios; the evaluation set is 6 unseen buildings $\times$ 5 unseen stations, giving 30 scored rollouts per submission. The simulator is deterministic, so a single rollout per pair carries no sampling noise and repeated rollouts would add nothing (\ref{sec:metrics}); breadth across buildings and climates, rather than repetition, is what supports conclusions about transfer.

\paragraph{Ground-truth confidentiality.}
The 6 held-out building configurations are generated from random seeds known only to the Organizers and stored in a private repository. No artifact derived from these buildings — floorplans, AHU divisions, baseline returns — is released until after the competition closes.

\paragraph{Feature leakage.}
Three mechanisms address leakage: (i) the train/evaluation split is across building \emph{configurations} rather than across time within a single building, so participants cannot memorize per-building physics; (ii) evaluation buildings are generated from seeds withheld from participants and never appear in any released artifact prior to the competition close; (iii) the policy receives the action spec, floorplan, and AHU divisions of each evaluation building only at runtime inside the evaluation sandbox, so they cannot be inspected, indexed, or used to pre-compile per-building heuristics during training.

\subsection{Tasks and application scenarios}\label{sec:tasks}
\paragraph{Real-world scenario.}
This task mirrors managing a diverse real-estate portfolio where per-building tuning is too costly and risky. A successful policy acts as a universal controller that adapts to unseen buildings using only floorplans and HVAC inventory, enabling scalable, portfolio-wide building optimization.

\paragraph{Scientific challenge.}
Varying building dynamics prevent overfitting, yet shared physics and established baselines (\ref{sec:baselines}) make the problem tractable. The core hurdle is transferability: participants must learn a single controller that successfully conditions its decisions on the specific building it is operating: a foundation model for smart buildings. An open question that we also hope to explore is, whether domain specific models will win out\cite{goldfeder2026ai}, or whether generic foundation models will generalize best.

\paragraph{Ethics.}
The competition uses purely simulated environments with no human subjects or personal data. Post-contest live trials follow strict institutional safety protocols, ensuring participants never directly control physical equipment. The project aims solely to reduce carbon emissions and poses no foreseeable societal harm.

\subsubsection*{Action space}
Each environment has a hybrid action with discrete and continuous parts. Every AHU and the Hot Water System (HWS) has one discrete \emph{run command} (on/off). Each AHU has three continuous setpoints: supply-air temperature, static pressure (duct airflow), and recirculation fraction. The HWS has two: supply-water temperature and pump differential pressure. Each VAV box (one per zone) has two: damper position and reheat valve position. Since an AHU sends one supply-air temperature to all its zones, the dampers and reheat valves are the only per-zone levers. Discrete commands are in $\{0,1\}$; continuous setpoints are normalized to $[-1,1]$. For a building with $K$ AHUs and $V$ zones, the action is a dict with \texttt{discrete\_action} of shape $(K{+}1,)$ and \texttt{continuous\_action} of shape $(3K{+}2{+}2V,)$. The simulator enforces ventilation floors (minimum static pressure for a running AHU, at most 70\% recirculation, dampers at least 10\% open), so shutting all air off is not an energy strategy. Per-index mappings are given in the building metadata.

\subsubsection*{Observation Space}
Each environment returns a flat observation vector whose length depends on the building. It contains, per zone, air temperature, heating/cooling setpoints, damper and reheat positions, and airflow setpoint; per AHU, supply-air temperature, static pressure, airflow, recirculation, and fan status; the HWS supply-water temperature; building occupancy and comfort-schedule flags; outside air temperature; and time of day and day of week. The policy also receives building metadata: action indices, observation field names, the AHU-to-zone assignment, floor plan, schedule, and normalization constants.

For each unseen building in the evaluation set, the floorplan, AHU divisions, and action spec are delivered to the policy at runtime as part of the observation (see \ref{sec:data}).

\subsubsection*{Policy API}
Contestants submit a Python class with two methods, subclassing the policy base class supplied by the evaluation gateway, and hand that class to the inference server:

\begin{lstlisting}[language=Python,basicstyle=\ttfamily\footnotesize]
from kaggle_evaluation import smart_buildings_inference_server

class MyPolicy(smart_buildings_inference_server.Policy):
    def __init__(self, building_meta_data, action_spec):
        super().__init__(building_meta_data, action_spec)
        ...
    def get_action(self, obs) -> dict: ...

smart_buildings_inference_server.SmartBuildingsInferenceServer(MyPolicy).run()
\end{lstlisting}

\noindent The constructor is called once per building with its metadata (floor plan, AHU-to-zone mapping, action and observation indices) and \texttt{action\_spec} (keys, shapes, and ranges). \texttt{get\_action} maps the current observation to an action dict matching \texttt{action\_spec} exactly. The class name is the participant's own choice; only the two signatures are fixed. Submissions take the form of a Kaggle notebook; the full submission instructions and the current template notebook are given on the Kaggle competition submission page.

\subsection{Metrics}\label{sec:metrics}

Evaluation uses the regret-based reward, at each 5-minute step $t$,
\[
r_t = w_p\, p_t - w_c\, c_t - w_e\, e_t,
\qquad (w_p, w_c, w_e) = (0.6,\ 0.2,\ 0.2).
\]
The reward is a weighted sum of three terms: $p_t$ is occupant comfort (productivity), $c_t$ is the energy cost (electricity and gas), and $e_t$ is the carbon emission. The comfort term drops as zone temperatures leave the comfort band; the drop is linear up to 5\,K outside the band and quadratic and unbounded beyond, so abandoning comfort is never free. It is summed over zones weighted by each zone's own occupancy rather than its floor area, so a single badly conditioned zone can dominate an episode. Weights are fixed by the Organizers and identical for all submissions. Each scenario runs for 30 days at a 5-minute timestep and every step is scored, so the episodic return is $R^{(s)}(\pi) = \sum_{t=1}^{T} r_t^{(s)}$ with $T = 8640$.

\paragraph{The reward is bounded above by zero.} All three terms are penalties. The comfort term $p_t$ is a \emph{regret}: it is $0$ when every occupied zone sits inside its comfort band and falls negative otherwise, without bound. Energy cost $c_t$ and carbon emission $e_t$ are non-negative quantities entering with a negative sign, since the building cannot export energy. Hence $r_t \le 0$ always, with equality only when comfort is perfect and the plant draws nothing. Measured over a full $8640$-step episode, the largest per-step reward observed is exactly $0$, attained at $95$ steps, all of them with the building unoccupied and the plant idle.

\paragraph{Leaderboard score.} Let $\bar{R}^{(s)}(\pi) = R^{(s)}(\pi)/T$ be the mean per-step reward on scenario $s$. The leaderboard metric is the mean absolute error of these values against a target of zero,
\[
\mathrm{MAE}(\pi) \;=\; \frac{1}{|\mathcal{S}|}\sum_{s\in\mathcal{S}} \bigl|\bar{R}^{(s)}(\pi) - 0\bigr| \;=\; -\frac{1}{|\mathcal{S}|}\sum_{s\in\mathcal{S}} \bar{R}^{(s)}(\pi),
\]
where $\mathcal{S}$ is the set of evaluation scenarios. \textbf{Lower is better.} The second equality holds because $r_t \le 0$, so the metric is simply the mean magnitude of the per-step loss. The target of zero is not an arbitrary reference point: it denotes perfect comfort delivered at no energy and no carbon cost, which is the unattainable ideal of the control problem. Because the transformation is monotone, minimizing $\mathrm{MAE}$ is identical to maximizing the return on every individual scenario.

A submitted policy is evaluated on each of the 6 held-out buildings at each of 5 weather stations, giving 30 scenarios. The simulator is deterministic: the same policy on the same scenario reproduces the same return bit for bit, across processes and machines, so each (building, weather) pair is run exactly once and no averaging over repeated rollouts is required. The headline leaderboard score is the mean over all 30 scenarios, with no grouping by building or weather.

The Organizers re-evaluate the top 7 submitted policies end-to-end at a fixed simulator commit before announcing final standings; because the simulator is deterministic, this re-run verifies rule compliance rather than estimating variance.

\subsection{Baselines}\label{sec:baselines}

\paragraph{Baselines.}
Building upon earlier work\cite{goldfeder2025evaluating}, we ship four hand-written reference controllers, all of which run the plant: \textbf{Bang-Bang} (\texttt{bang\_bang}) switches each AHU on or off around the comfort band; \textbf{Fixed Setpoint} (\texttt{fixed\_setpoint}) holds one comfort band around the clock with no night setback; \textbf{PID} (\texttt{pid}) uses proportional control with fan pressure scaled by control effort; and \textbf{PID + Water} (\texttt{pid\_water}) adds modulating reheat and a demand-driven hot-water reset. Each is a controller a practitioner might plausibly write. An \textbf{All Off} policy is shipped as an additional reference point but is not counted among the four, because it does not run the plant at all. The baselines do not enter the leaderboard metric, which needs no reference policies; they give participants a published performance range to develop against, and their returns on the public training scenarios are shipped with the starter kit. On those scenarios \texttt{bang\_bang} sets the lower end of the range, while the upper end is set by \texttt{pid\_water} or, in hot and humid weather, by \texttt{fixed\_setpoint}.

\begin{figure}[ht]
    \centering
    \includegraphics[width=\linewidth]{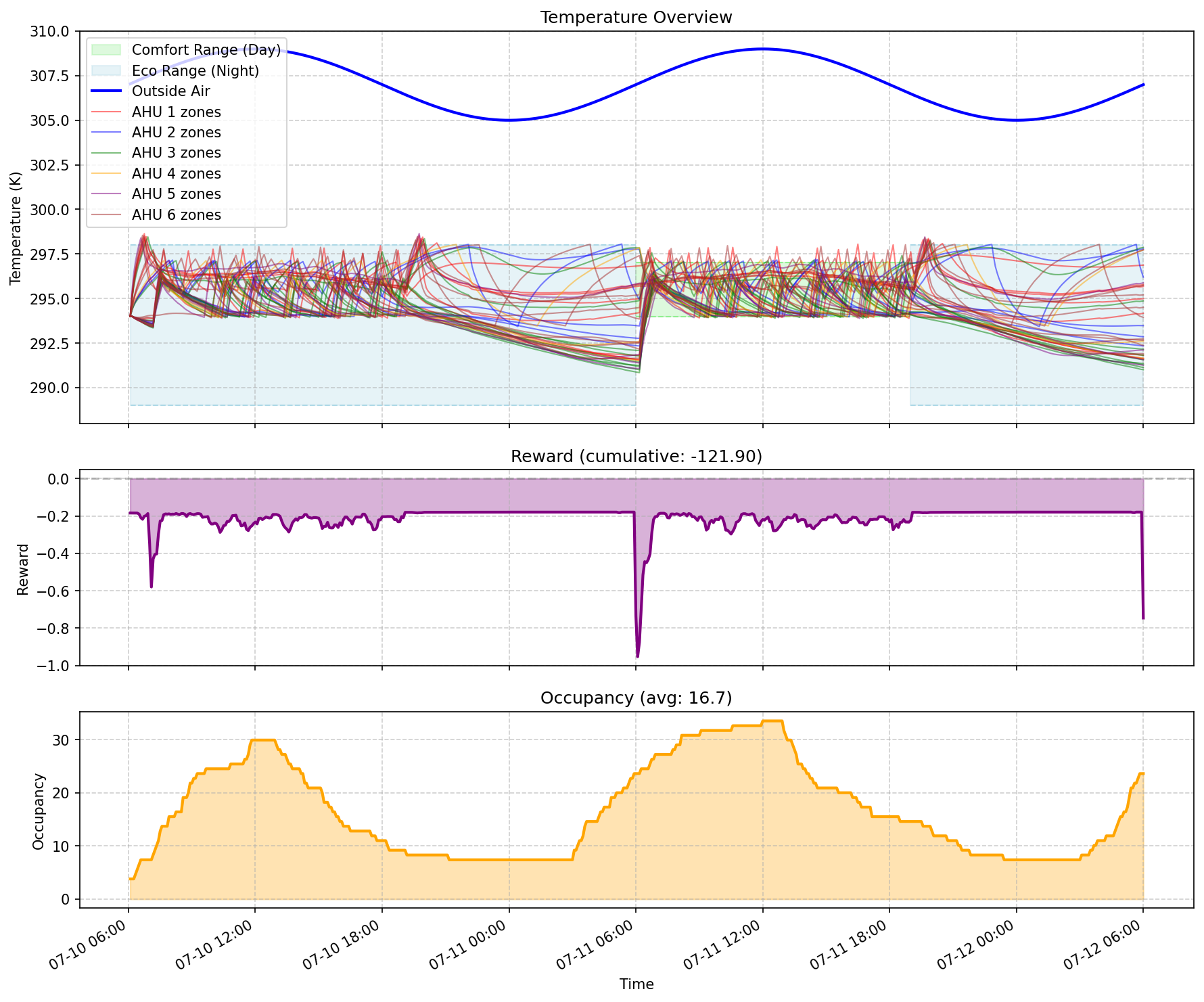}
    \caption{One rollout: zone temperatures against the comfort band, per-step reward, and occupancy.}
    \label{fig:example_plot}
\end{figure}

Each scenario runs for 30 days, all of which are scored. No single baseline is best everywhere, which confirms the transfer challenge. For each policy on each scenario we produce the overall return and a plot charting zone temperature, return, occupancy, and reward-component contributions over time (Figure~\ref{fig:example_plot}); a video rendering of the building over the scenario can be produced on request.

\section{Organizational aspects}

\subsection{Protocol}\label{sec:protocol}
The competition will be hosted on Kaggle, with the prize pool jointly sponsored by Kajima, Distyl, and KODE Labs. These will be collectively referred to as Organizers from here on. A Participant or group of Participants (a ``Team'', up to four members) submits an Entry consisting of (i) a Python module implementing the \texttt{TransferPolicy} interface, (ii) the model weights or other artifacts needed to instantiate it, and (iii) a reproducibility bundle (code archive, seed list, dependency manifest). The latest valid Submission per Team is the official Entry. Submissions are sandboxed and executed by the Organizers against the held-out 6-building evaluation suite via the \texttt{evaluate.py} interface.

\paragraph{Two-phase structure.}
The competition runs in two phases. During the \textbf{development phase}, the Kaggle public leaderboard scores submissions against a rotating, publicly-visible subset of evaluation (building, weather) pairs; this gives participants signal without exposing the full held-out set. During the \textbf{final phase}, all submitted policies are re-evaluated against the complete held-out set (\ref{sec:metrics}), and final standings are determined by the leaderboard metric of \ref{sec:metrics}, mean absolute error against zero, on the private leaderboard, where lower is better. The public/private split mitigates overfitting to the visible evaluation signal.

\paragraph{Anti-cheating and fairness.}
\begin{enumerate}[leftmargin=*]
    \item The held-out 30 scenarios are never released to participants, in any form, before the competition closes.
    \item Policies are executed in an isolated container at evaluation time with no network access, blocking external lookup tables or remote inference.
    \item Per-rollout wall-clock is bounded at $10\times$ the bang-bang baseline runtime so solutions cannot succeed by unbounded online planning.
    \item Submitted code archives are scanned for hard-coded building IDs or sensor signatures before scoring; matches are grounds for disqualification.
\end{enumerate}


\paragraph{Evaluation harness.}
Participants develop against \texttt{evaluate.py}, the harness shipped in the starter kit: a set of (building, weather) configs is specified and each is rolled out once for $T = 8640$ steps, after which \texttt{score\_leaderboard.py} reports the leaderboard metric of \ref{sec:metrics} together with the shipped baseline returns for comparison, so the four baselines need not be re-run. Scoring on the leaderboard is performed instead by an inference-server gateway, which owns the simulator, iterates the held-out scenarios and queries the submitted policy once per step; a submission therefore contains no scenario list and no rollout loop. Reference policies (the four baselines from \ref{sec:baselines}), the public training scenarios with their baseline returns, and example invocations are shipped in the starter kit (\url{https://github.com/smart-buildings-2026/starter-kit}).

\subsection{Rules and Engagement}\label{sec:rules_and_engagement}

The Challenge will be run in accordance with the rules and regulations posted on the Kaggle page.

\subsection{Schedule and readiness}
Our competition preparation and execution timeline is outlined below.

\begin{center}\small
\begin{tabular}{ll}
\toprule
\textbf{Date} & \textbf{Task / Event} \\
\midrule
Nov 13, 2026  & Round One deadline (NeurIPS poster; \$2K in prizes) \\
Dec 13, 2026  & Paris Workshop \\
Jan 15, 2027  & Final deadline (\$38K in prizes) \\
\bottomrule
\end{tabular}
\end{center}


\subsection{Competition promotion and incentives}\label{sec:promotion}

\paragraph{Incentives.}
The \$40{,}000 total prize pool will be split across podium positions on the private leaderboard. Beyond monetary awards, top teams will receive: (i) an invited podium talk at the NeurIPS 2026 Competition Track workshop; (ii) co-authorship on the consortium write-up of the competition results (named-author status for teams contributing a methods description that meets the standard threshold; consortium-author status otherwise); (iii) invitation to the post-contest live-trial program at the four physical partner sites.

\bibliographystyle{plainnat}
\bibliography{references.bib}
\appendix

\end{document}